\pdfoutput=1

\documentclass[11pt]{article}

\usepackage[table]{xcolor}
\usepackage[preprint]{coling}
\usepackage{amsmath}
\usepackage{tabularx}
\usepackage{algorithm}
\usepackage{algpseudocode}
\usepackage{hyperref}
\usepackage{booktabs}
\usepackage{nomencl}
\usepackage{times}
\usepackage{multirow}
\usepackage{latexsym}
\usepackage{orcidlink}

\usepackage[utf8]{inputenc}

\usepackage{microtype}
\usepackage{inconsolata}
\usepackage{graphicx}
\usepackage{pgfplots}
\usepackage{pgfplotstable}
\usepackage{tikz}
\usepackage{amsmath}
\usepgfplotslibrary{polar}
\pgfplotsset{compat=1.17}

\title{Contextual Breach: Assessing the Robustness of\\ Transformer-based QA Models}

\author{
Asir Saadat\textsuperscript{*}\\
\textit{Rochester Institute of Technology}\\
\texttt{as5606@g.rit.edu}\\
\And
Nahian Ibn Asad\textsuperscript{*}\\
\textit{Islamic University of Technology}\\
\texttt{nahian36@iut-dhaka.edu} }

\begin{document}
\maketitle
\def\thefootnote{*}\footnotetext{These authors contributed equally to this work.}
\begin{abstract}

Contextual question-answering models are susceptible to adversarial perturbations to input context, commonly observed in real-world scenarios. These adversarial noises are designed to degrade the performance of the model by distorting the textual input. We introduce a unique dataset that incorporates seven distinct types of adversarial noise into the context, each applied at five different intensity levels on the SQuAD dataset. To quantify the robustness, we utilize robustness metrics providing a standardized measure for assessing model performance across varying noise types and levels. Experiments on transformer-based question-answering models reveal robustness vulnerabilities and important insights into the model's performance in realistic textual input. 
\end{abstract}

\section{Introduction}


The capacity to respond to user inquiries in natural language is enhanced by contextual question-answering (QA) models, which use the context of a query to produce relevant responses.  Transformer-based models \cite{vaswani2017attention} models like the pre-trained encoder BERT \cite{devlin2018bert} remain widely used in contextual QA tasks. These models remain relevant due to being open-source, infers fast, suitable for tasks demanding efficiency \cite{subendhu2021continual}, can be easily fine-tuned \cite{paul2023fine,imran2024leveraging}, and suitable for domain-specific datasets \cite{lee2020biobert, alsentzer2019publicly}.



Adversarial noise \cite{szegedy2013intriguing, ebrahimi2017hotflip} refers to the deliberate perturbations or distortions into the input data to deceive and disrupt the performance of machine learning models, including QA systems. These perturbations can lead to a range of issues, including misinterpretation of the query, retrieval of irrelevant information, and generation of incorrect answers 
 \cite{jia2017adversarial, li2020bert, jin2019bert}. Researchers have explored various methods to generate adversarial examples that specifically target the question formulation process \cite{jia2017adversarial, hwang2019comprehensive}. 

While significant strides have been made in adversarial attacks targeting questions \cite{jia2017adversarial, ribeiro2018semantically, wang2019does} within QA models, there exists a notable gap in exploring adversarial noise added specifically to the context passages. However, systematic exploration of how varying levels of adversarial noise injected into context passages affect the performance of the model remains underexplored. 

This gap is further enhanced by the absence of a proper measure. F1-score and exact matching are highly sensitive to minor variations in the predicted answers. In a noisy context, if the model captures the essence of the correct answer but has slight differences (e.g., synonyms, word order), these metrics could penalize the model harshly. In noisy conditions, it is important to evaluate how close the answer of a model is to the correct one, even if it is not an exact match. There is this lack of consistency in the approaches used to evaluate robustness, necessitating frameworks that evaluate performance over a range of perturbation kinds and intensities. The lack of a numerical measure for robust assessment that successfully combines these findings is a significant challenge.

As QA systems became more sophisticated, the need to evaluate and enhance their robustness emerged. Robustness in QA models refers to their ability to maintain performance when faced with noisy, incomplete, or adversarial inputs. This aspect is crucial for deploying QA systems in real-world applications where inputs can be unpredictable and varied. Addressing the above issues, we propose a novel framework for evaluating a QA model’s robustness to different contextual corruptions. Our contributions can be summarized as:

\begin{enumerate}
        \item We introduced the first benchmark for robustness evaluation of contextual QA models comprising of 30,000 QA pairs with adversarial contexts. 

        \item  We utilized $3$ robustness evaluation metrics to assess the contextual robustness of QA models against adversarial noise injected into the context. These metrics, unlike conventional ones, are uniquely tailored to capture the nuances of adversarial perturbations, addressing gaps where traditional evaluation methods fall short.
            
        \item We evaluated $5$ transformer-based QA models on $7$ contextual corruption functions at $5$ intensity levels.
\end{enumerate}

\section{Related Work}

\subsection{Robustness in Contextual QA}

  
\citet{jia2017adversarial} showed that many reading comprehension systems can be easily fooled by superficial changes by adding distractor sentences. \citet{rajpurkar2018know} includes unanswerable questions that require the model to identify when no answer is available in the context. \citet{jin2020bert} present "TEXTFOOLER", a straightforward but effective baseline to produce hostile text, and used it for textual entailment and text classification. However, the primary objective of these studies was to deceive the models, resulting in adversarial noise that was not representative of real-world examples. Additionally, the adversarial noise used in these works lacked diversity. Their focus was solely on creating adversarial inputs to mislead the models, leading to limited variations and an insufficient range of adversarial noise types.



\subsection{Sentence-Level Adversarial Noise}
Previous works, such as \citet{jia2017adversarial, liu2020robust, wang2018robust}, have explored adding adversarial noise at the sentence level to assess robustness, focusing primarily on introducing misleading text to test the model’s ability to maintain accuracy. While these studies effectively evaluate model resilience to sentence-level perturbations, they provide a limited evaluation of robustness. \citet{iyyer2018adversarial, belinkov2017synthetic} investigated adversarial example generation through syntactic paraphrasing but did not assess QA model performance. 

To gain a better understanding of model robustness, it is crucial to extend evaluation beyond sentence-level adversarial noise \emph{i.e.} including character-level and word-level perturbations to capture a fuller range of adversarial scenarios. \citet{ebrahimi2017hotflip} proposed a method for generating adversarial examples for text classification tasks which included character-level flips. \citet{pruthi2019combating} introduced misspelling words in a targeted manner to degrade model performance which included insertion, deletion, and substitution operations. 



\subsection{Assessment Criteria}
Present-day techniques for analyzing the robustness of QA models frequently depend on data perturbations and accuracy variations; success rate and conventional evaluation metrics, such as precision, recall, and F1-score. The usage of F1-score and exact matching as accuracy do not properly justify the robustness of any model. Several metrics, such as Robustness Index \cite{moritz2016robustness} and Error Rate \cite{farhan2024visual} were introduced as robustness assessment metrics.

\citet{gan2019improving} evaluates the robustness of QA models in response to question paraphrasing. \citet{antol2015vqa} rely heavily on F1-score and Exact Match to compare different models’ ability to match ground truth answers. \cite{bhandari2024robustness} emphasizes their capability to reliably process tabular data. 
 \citet{jia2017adversarial} demonstrated that adversarial examples could drastically reduce exact match and F1 scores in QA models, even though the models provided answers that were semantically correct but not textually identical. These metrics alone do not provide a holistic view of a model’s robustness or generalization capabilities across different input variations. 

\begin{figure*}[ht]
  \includegraphics{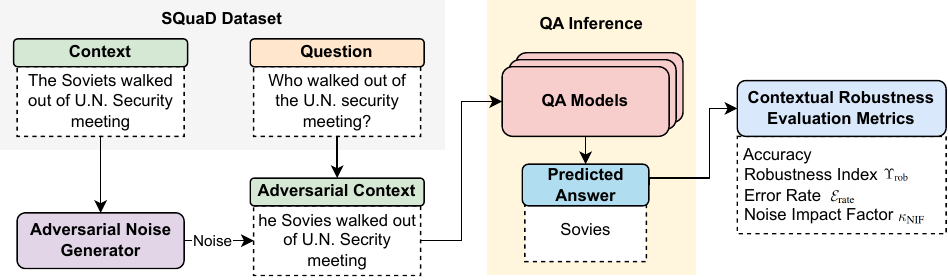}
  \caption{Overview of the Robustness Evaluation Framework for QA Models. The figure illustrates the process of adding adversarial noise to the context from the SQuAD dataset and feeding the perturbed context into QA models for inference. The predicted answers are then evaluated using contextual robustness metrics such as Accuracy, Robustness Index, Error Rate, and Noise Impact Factor.}
  \label{fig:Robustness Index}
\end{figure*}

\section{Methodology}

\makenomenclature

\label{appendix:nomenclature}
\renewcommand\nomgroup[1]{%
  \item[\bfseries
  \ifstrequal{#1}{A}{Core Components}{%
  \ifstrequal{#1}{B}{Perturbation Functions}{%
  \ifstrequal{#1}{C}{Context and Noise}{%
  \ifstrequal{#1}{D}{Random Selection}{%
  \ifstrequal{#1}{E}{Noise Severity}{%
  \ifstrequal{#1}{F}{Character and Word-Level Operations}{%
  \ifstrequal{#1}{G}{Special Conditions and Checks}{%
  \ifstrequal{#1}{H}{Robustness Metrics}{}}}}}}}}%
]}
\setlength{\nomitemsep}{-\parskip}

\nomenclature[A, 05]{\( S(P) \)}{The set of sentences in the paragraph \( P \)}
\nomenclature[A, 06]{\( W(S) \)}{The set of words in a sentence \( S \)}
\nomenclature[A, 07]{\( \vert W(S)\vert \)}{The number of words in the sentence \( S \)}
\nomenclature[B, 06]{\(D(w)\)}{The deletion of a single character from the word \(w\), chosen randomly}
\nomenclature[B, 08]{\(A(w)\)}{Function that adds a single random character to the word \(w\) or a space}
\nomenclature[B, 08]{\(T(w)\)}{Function that adds a random word \(w\)}
\nomenclature[B, 09]{\(N(w)\)}{A neighboring character of \(w\) on the keyboard}
\nomenclature[B, 12]{\(\mathcal{X}_i, \mathcal{X}_j\)}{Randomly selected indices of words in the sentence to be swapped}
\nomenclature[B, 13]{$\text{swap}(\cdot)$}{Function that swaps two words \(w_i\) and \(w_j\)}
\nomenclature[B, 14]{\ensuremath{\mathcal{I}_j}}{Randomly selects an index \(\mathcal{I}_j\) from the range of valid indices of \(w_j\)}
\nomenclature[B, 16]{\( w' \)}{The word after character perturbation}
\nomenclature[B, 17]{\( P' \)}{The modified paragraph after character deletions}

\nomenclature[C, 18]{\( \text{Ctx}_0 \)}{Nominal context (without noise)}
\nomenclature[C, 19]{\( \text{Ctx}_{noise,L} \)}{Noisy context at level \(L\)}

\nomenclature[D, 01]{\( w_j \)}{Randomly selected word \(w\) from sentence \(S\)}
\nomenclature[D, 03]{\(\mathcal{R}_j(\cdot)\)}{A random selection of the \(j\)-th word}
\nomenclature[D, 05]{\(c\)}{A random character chosen from the set of possible characters or an empty space}

\nomenclature[E, 01]{\( L \)}{The severity level for character deletion}
\nomenclature[E, 02]{\( r(L) \)}{The severity ratio, defined as \( \frac{L}{5} \)}
\nomenclature[E, 05]{\(n\)}{Number of perturbations defined by \(\text{int}(\vert W(S)\vert \times r(L))\)}

\nomenclature[F, 15]{\( w[i] \)}{The \(i\)-th character in the word \(w\)}
\nomenclature[F, 11]{\(\mathcal{G}_M(S)\)}{Transformation of sentence \(S\) under the randomly chosen grammatical mistake \(M\)}

\nomenclature[G, 1]{\(\mathcal{P}(w)\)}{A predicate that returns true if \(w\) is a proper noun}
\nomenclature[G, 2]{\(\text{Syn}(w)\)}{A function that returns a synonym of the word \(w_j\)}
\nomenclature[G, 3]{\(\texttt{syn\_avail}\)}{A condition that checks whether synonyms for \(w_j\) are available}
\nomenclature[H, 1]{\( \Upsilon_{\text{rob}} \)}{Robustness Index}
\nomenclature[H, 2]{\(\mathcal{E}_{\text{rate}}\)}{Error Rate}
\nomenclature[H, 3]{\( \kappa_{\text{NIF}} \)}{Noise Impact Factor}

\printnomenclature

\subsection{Contextual QA}
We define the question answering model $f$, trained on parameters $\theta$, such that $f(C,Q) \rightarrow Y$, where $C$ is the context, $Q$ is the input question and $Y$ is the predicted answer based on the given context. Answer generation can be formulated as:
\begin{equation}
    f(c,q) = \operatorname{argmax} P(y|C,Q;\theta)
\end{equation}

\subsection{Adversarial Noise Function}
Adversarial noise functions are defined as transformation functions that apply textual perturbations commonly observed in real-life. We categorize these perturbations for several scales \emph{e.g.} characters, words, sentences, and semantics. While the overall meaning should remain the same, the replacement can sometimes lead to subtle differences in connotation or usage. The key notion behind introducing the perturbations is to model realistic textual perturbations observed in practical scenarios. 

\subsubsection*{Synonym Replacement}
Substitutes with their synonyms to test the model’s ability to understand lexical variations which tests the model's ability to understand meaning beyond specific word choices, ensuring robustness to lexical variation.



\[
Syn(\mathcal{R}_j(W(S))) = 
\begin{cases}
    w_j, & \text{if } \mathcal{P}(w_j), \\
    Syn(w_j), & \text{if } \texttt{syn\_avail}
\end{cases}
\]

\subsubsection*{Character Deletion}

Randomly removes characters from words, leading to misspellings and potentially making words unrecognizable, evaluating the model’s tolerance to minor misspellings. 




\[
D(\mathcal{R}_j(W(S))) = w_j[:\mathcal{I}_j] + w_j[\mathcal{I}_j + 1:]
\]


\subsubsection*{Character Insertion}
Adds random characters to words, causing misspellings and making the text more difficult to read. 


\[
A(\mathcal{R}_j(W(S))) = w_j[:\mathcal{I}_j] + c + w_j[\mathcal{I}_j:]
\]

\subsubsection*{Word Swapping}
Disrupts the natural order of words in a sentence evaluating the model's reliance on word order.

\begin{align*}
P' &= \sum_{S \in S(P)} \Bigg[ \text{Join}\left(\left\{ w_k' \mid w_k' = \text{swap}(w_{\mathcal{X}_1}, w_{\mathcal{X}_2}) \right.\right. \\
&\left.\left. \text{ for } k \in \{1, \ldots, n\} \right\} \right) \Bigg]
\end{align*}

\subsubsection*{Random Insertion}
Involves inserting random words into the text, which can distract from the main information and confuse the reader that tests the model’s ability to filter out irrelevant information and maintain focus on context.


\begin{align*}
P' &= \sum_{S \in S(P)} \Bigg[ \text{Join}\Big(W(S) \cup \\
& \quad \left\{\mathcal{T}(\mathcal{W})_1, \ldots, \mathcal{T}(\mathcal{W})_n \right\}\Big)\Bigg]
\end{align*}

\subsubsection*{Typing Mistake}
Mimics common typographical errors, such as those that occur from fast or inaccurate typing that simulates human error in text entry.

\[
w_j' = 
\begin{cases}
N(w_j[I_j]), & p = 0.5 \\
\text{swap}(w_j[I_j], w_j[I_j+1]), & p = 0.5 \\
\end{cases}
\]

\subsubsection*{Grammatical Mistake}
Involves introducing grammatical errors like verb tenses, subject-verb agreement, preposition, article, double negation and misplaced modifier at the sentence level analyzing the model's ability to handle ungrammatical input and still comprehend meaning.

 \[
P' = \sum_{S \in S(P)} \left[ \mathcal{G}_{\mathcal{R}(\mathcal{M})}(S) \right]
\]



\section{Evaluation Metrics} 
Metrics that evaluate how performance changes with increasing noise offer a more nuanced view of robustness are more suitable.

\subsection{Accuracy of Question Answering Models}
In evaluating the correctness of answers produced by a QA model, cosine similarity serves as an effective metric for understanding the semantic alignment between the predicted and referenced answers. Unlike binary metrics that solely focus on exact matches, cosine similarity measures the cosine of the angle between two vectors representing the text embedding of the answers. 


\[
\text{Accuracy} = cosine(\textbf(A), \textbf(B))
\]
where \(\mathbf{A}\) and \(\mathbf{B}\) are the vector representations of the predicted and referenced answers, respectively.

\subsection{Robustness Index} Quantifies the variation of model performance when the nominal (or baseline) conditions are perturbed or altered. 
\begin{equation*}
 \Upsilon_{\text{rob}}  = \frac{1}{ L } \sum_{i=1}^L \frac{|P_{\text{nominal}} - P_{\text{perturbed}}(i)|}{P_{\text{nominal}}}
\end{equation*}

The formula directly measures how sensitive the performance of a model with respect to the changes in the input data. By calculating the difference between the performance on nominal data $P_{\text{nominal}}$ and on perturbed data $P_{\text{perturbed}}(i)$, it captures the extent to which the model's output changes when the input is altered. This difference is critical for understanding the robustness of the model.

\subsection{Error Rate}
In determining the robustness of a model to varying levels of noise, we examine the relationship between the noise levels \(x_i\) and the corresponding model accuracy \(y_i\). This relationship is quantified by fitting a linear regression model, where the slope of the best-fit line is given by:

\[
\mathcal{E}_{\text{rate}} = \frac{\sum_{i=1}^{n} (x_i - \bar{x})(y_i - \bar{y})}{\sum_{i=1}^{n} (x_i - \bar{x})^2}
\]

The error rate signifies the rate at which the accuracy of the model declines as noise increases, providing a comprehensive overview of the performance degradation. A smaller negative slope indicates that accuracy is less affected by noise, reflecting greater robustness. In contrast, a steeper slope implies rapid accuracy deterioration with noise, signifying reduced robustness. 




\subsection{Noise Impact Factor }
The Noise Impact Factor (NIF) is defined to quantify the effect of noise on both the context and the resulting answers in a model. It is given by:

\[
\kappa_{\text{NIF}} = \frac{1}{L} \sum_{i=1}^{L} \frac{\text{Accuracy}_{noise,i}}{cosine(\text{Ctx}_0, \text{Ctx}_{noise,i})}
\]

The NIF ensures that the metric does not just reward high accuracy, but also considers how challenging the noisy input was compared to the original. This metric is distinct because it considers the perturbed context during evaluation. A higher NIF occurs when the accuracy remains high despite the similarity score being low. This means the model performs well even when the input data has been significantly altered, suggesting strong robustness.

\section{Experimental Setup}

\subsection{Hardware and Implementation Details}
In our experiments, we utilized an NVIDIA GeForce GTX 1650 GPU, equipped with 4GB of VRAM, to efficiently train and evaluate the models. The models were primarily sourced from the Hugging Face Transformers library \cite{wolf2020transformers}, which provided a diverse range of pre-trained models tailored for various NLP tasks, including Question Answering (QA).

\subsection{Dataset}
We have developed a dataset derived from the SQuAD dataset \cite{rajpurkar2016squad}, encompassing the first 30,000 question-answer pairs associated with 6,770 unique contexts. To enhance the variability of this dataset, adversarial noises were systematically introduced into the contexts. Specifically, we utilized 7 distinct types of noise, each applied at five varying levels of severity, with level 1 representing the lowest severity and level 5 the highest. The severity levels were increased by adjusting the number of perturbations which is determined by \(\text{int}(\vert W(S)\vert \times r(L)) \). This methodology produced a diverse range of unique combinations of noise type and severity level, consistently applied across 30,000 question-answer pairs.


\subsection{Evaluated Models}
We leveraged a diverse set of models including BERT \cite{devlin2018bert}, DeBERTa\cite{he2020deberta}, ELECTRA\cite{clark2020electra}, DistilBERT\cite{sanh2019distilbert}, and RoBERTa \cite{liu2019roberta} \footnote{Models were downloaded from \url{https://huggingface.co/} }. All five models harness the capabilities of transformer architectures \cite{vaswani2017attention} to excel in language processing tasks.

\definecolor{mygrey}{gray}{0.93}
\definecolor{myred}{rgb}{1.0, 0.9, 0.9}
\definecolor{mygreen}{rgb}{0.9, 1.0, 0.9}
\definecolor{myblue}{rgb}{0.7, 0.9, 1.0}
\definecolor{mysalmon}{rgb}{1.0, 0.8, 0.7}
\definecolor{mylavender}{rgb}{0.902, 0.902, 0.980}
\definecolor{myyellow}{rgb}{0.980, 0.980, 0.824}


\begin{table*}[ht]
\label{tab:errorVRE}
\centering

\resizebox{\textwidth}{!}{%
\begin{tabular}{c|c|ccccccc}

\hline
\textbf{Model} & \textbf{Lvl} &    \textbf{Char Del}       & \textbf{Gram Err} & \textbf{Char Ins}    & \textbf{Word Ins}  & \textbf{Word ReOrd}  & \textbf{Syn Repl}      & \textbf{Typo}            \\ \hline
\multirow{6}{*}{BERT}  & 0                    & \multicolumn{7}{c}{\cellcolor{mygrey} \textbf{0.765}}                                                                                                                                                                                                                                      \\ \cline{3-9} 
                       & 1                    &  0.683      & 0.731 & 0.684 & 0.654 & 0.628 & 0.729 & 0.691           \\
                       & 2                    & 0.623       & 0.710 & 0.630 & 0.590 & 0.580 & 0.698 & 0.632   \\
                       & 3                    & 0.584       & 0.694 & 0.591 & 0.551 & 0.561 & 0.670 & 0.593  \\
                       & 4                    & 0.556 & 0.680 & 0.559 & 0.524 & 0.555 & 0.643       & 0.561           \\
                       & 5                    &0.535 & 0.668 & 0.535&{\cellcolor{myred}\textbf{0.504}} & 0.551& 0.620   & 0.538  \\ \hline
\multirow{6}{*}{DeBERTa}  & 0                    & \multicolumn{7}{c}{\cellcolor{mygrey} \textbf{0.993}}                                                                                                                                                                                                                                      \\ \cline{3-9} 
                       & 1                    & 0.933      & 0.970 & 0.947 & 0.960 & 0.855 & 0.971 & 0.941        \\
                       & 2                    & 0.886       & 0.954 & 0.910 & 0.928 & 0.801 & 0.951 & 0.898      \\
                       & 3                    & 0.846       & 0.943 & 0.877 & 0.896 & 0.784 & 0.932 & 0.862         \\
                       & 4                    & 0.815 & 0.932 & 0.846 & 0.864 & 0.776 & 0.915       & 0.832            \\
                       & 5                    & 0.785 & 0.922 & 0.822 & 0.827 &\cellcolor{myred}\textbf{0.770} & 0.895       & 0.803   \\ \hline
\multirow{6}{*}{ELECTRA}   & 0                    & \multicolumn{7}{c}{\cellcolor{mygrey} \textbf{0.982}}                                                                                                                                                                                                                                      \\ \cline{3-9} 
                       & 1                    & 0.905       & 0.962 & 0.917 & 0.931 & 0.853 & 0.959 & 0.914            \\
                       & 2                    & 0.837       & 0.946 & 0.858 & 0.886 & 0.787 & 0.935 & 0.853             \\
                       & 3                    & 0.781 & 0.935 & 0.806 & 0.845 & 0.757 & 0.910       & 0.800            \\
                       & 4                    & 0.738 & 0.923 & 0.756 & 0.800 & 0.745 & 0.885       & 0.753            \\
                       & 5                    & \cellcolor{myred}\textbf{0.692} & 0.913 & 0.710 & 0.758 & 0.742 & 0.854       & 0.707            \\ \hline
                       
\multirow{6}{*}{DistilBERT}   & 0                    & \multicolumn{7}{c}{\cellcolor{mygrey} \textbf{0.999}}    

\\ \cline{3-9} 
                       & 1                    & 0.936       & 0.983 & 0.936 & 0.925 & 0.893 & 0.980 & 0.938            \\
                       & 2                    & 0.887       & 0.971 & 0.887 & 0.872 & 0.846 & 0.960 & 0.891             \\
                       & 3                    & 0.847 & 0.962 & 0.850 & 0.832 & 0.830 & 0.941       & 0.854            \\
                       & 4                    & 0.818 & 0.952 & 0.820 & 0.797 & 0.824 & 0.921       & 0.826            \\
                       & 5                    &  0.794 & 0.944 & 0.788 & \cellcolor{myred}\textbf{0.760} & 0.819 & 0.902       & 0.801            \\ \hline

\multirow{6}{*}{RoBERTa}   & 0                    & \multicolumn{7}{c}{\cellcolor{mygrey} \textbf{0.954}}

\\ \cline{3-9} 
                       & 1                    & {0.876}       & 0.923 & 0.892 & 0.882 & 0.776 & 0.920 & 0.882           \\
                       & 2                    & 0.811       & 0.907 & 0.834 & 0.803 & 0.682 & 0.892 & 0.825            \\
                       & 3                    & 0.752 & 0.892 & 0.778 & 0.716 & 0.641 & 0.866       & 0.770          \\
                       & 4                    & 0.701 & 0.880 & 0.721 & 0.631 & 0.625 & 0.834       & 0.715            \\
                       & 5                    & 0.646 & 0.867 & 0.663 & \cellcolor{myred}\textbf{0.563} & 0.619 & 0.795       & 0.666            \\ \hline
\end{tabular}%
}

\caption{Accuracy of the model across various severity levels for different adversarial noise. \colorbox{myred}{\textbf{Red}} indicate the minimum accuracy for a particular model.} 

\label{tab:noise-performance}

\end{table*}

\begin{table*}[ht]

\resizebox{\textwidth}{!}{%
\begin{tabular}{c|c|ccccccc}
\hline
\cellcolor{mygrey}\textbf{Model}&\cellcolor{mygrey}\textbf{Metrics} &\cellcolor{mygrey}\textbf{Char Del} &\cellcolor{mygrey}\textbf{Gram Err} &\cellcolor{mygrey}\textbf{Char Ins}    &\cellcolor{mygrey}\textbf{Word Ins}  &\cellcolor{mygrey}\textbf{Word ReOrd}  &\cellcolor{mygrey}\textbf{Syn Repl}      &\cellcolor{mygrey}\textbf{Typo}            \\ \hline
\multirow{3}{*}{BERT}                                                                                                                                
                       & \( \Upsilon_{\text{rob}} \textcolor{red}{\downarrow} \)                   & {\cellcolor{myred}\textbf{0.221}}      & {\cellcolor{myred}\textbf{0.090}}  & {\cellcolor{myred}\textbf{0.217}} & {\cellcolor{myred} \textbf{0.263}} & 0.249 & {\cellcolor{myred}\textbf{0.122}}  & {\cellcolor{myred}\textbf{0.212}}           \\
                       & \(\mathcal{E}_{\text{rate}} \textcolor{green}{\uparrow} \)                    & -0.045        & {\cellcolor{mysalmon} \textbf{-0.019}} & -0.045 & -0.050 & -0.037 & -0.029 & -0.045   \\
                       & \(\kappa_{\text{NIF}} \textcolor{green}{\uparrow} \)                    & {\cellcolor{myyellow} \textbf{4.148}}    & {\cellcolor{myyellow} \textbf{4.091}} & {\cellcolor{myyellow} \textbf{4.153}} & {\cellcolor{myyellow} \textbf{4.091}} & {\cellcolor{myyellow} \textbf{4.210}} & {\cellcolor{myyellow} \textbf{4.052}} & {\cellcolor{myyellow} \textbf{4.122}}  \\
                     \hline
\multirow{3}{*}{DeBERTa}                                                   
                       & \( \Upsilon_{\text{rob}} \textcolor{red}{\downarrow} \)                    & {\cellcolor{mygreen} \textbf{0.141}}      & 0.049  & {\cellcolor{mygreen} \textbf{0.113}}  & {\cellcolor{mygreen} \textbf{0.099}} & 0.197 & 0.061 & {\cellcolor{mygreen} \textbf{0.126}}        \\
                       & \(\mathcal{E}_{\text{rate}} \textcolor{green}{\uparrow} \)                    & {\cellcolor{myblue} \textbf{-0.040}}      & -0.013 & {\cellcolor{myblue} \textbf{-0.034}} & {\cellcolor{myblue} \textbf{-0.032}} & -0.039 & {\cellcolor{myblue} \textbf{-0.019}} & {\cellcolor{myblue} \textbf{-0.037}}     \\
                       & \(\kappa_{\text{NIF}} \textcolor{green}{\uparrow}\)                    & 5.803      & 5.480 & {\cellcolor{mylavender} \textbf{5.914}} & {\cellcolor{mylavender} \textbf{6.199}} & 5.743 & 5.518 & {\cellcolor{mylavender} \textbf{5.777}}         \\
                        \hline
\multirow{3}{*}{ELECTRA} 
                       & \( \Upsilon_{\text{rob}} \textcolor{red}{\downarrow} \)                    & 0.195       & 0.047 & 0.176 & 0.140 & 0.209 & 0.074 & 0.180            \\
                       & \(\mathcal{E}_{\text{rate}} \textcolor{green}{\uparrow} \)                    & -0.057       & -0.013 & -0.054 & -0.044 & -0.044 & -0.025 & -0.054             \\
                       & \(\kappa_{\text{NIF}} \textcolor{green}{\uparrow}\)                    & 5.488 & 5.431 & 5.567 & 5.925 & 5.627 & 5.410      & 5.477            \\
                       \hline
                       
\multirow{3}{*}{DistilBERT} 
                       &\( \Upsilon_{\text{rob}} \textcolor{red}{\downarrow} \)                    & 0.142       & {\cellcolor{mygreen} \textbf{0.036}} & 0.143 & 0.162 &{\cellcolor{mygreen} \textbf{0.156}} & {\cellcolor{mygreen} \textbf{0.058}} & 0.137           \\
                       & \(\mathcal{E}_{\text{rate}} \textcolor{green}{\uparrow} \)                    & {\cellcolor{myblue} \textbf{-0.040}}        & {\cellcolor{myblue} \textbf{-0.010}} & -0.041 & -0.046 & {\cellcolor{myblue} \textbf{-0.032}} & {\cellcolor{myblue} \textbf{-0.019}} & -0.038             \\
                       & \(\kappa_{\text{NIF}} \textcolor{green}{\uparrow}\)                    & {\cellcolor{mylavender} \textbf{5.821}} & {\cellcolor{mylavender} \textbf{5.568}} & 5.802 & 5.897 & {\cellcolor{mylavender} \textbf{6.000}} & {\cellcolor{mylavender} \textbf{5.564}}       & 5.751            \\
                        \hline

\multirow{3}{*}{RoBERTa}  
                       & \( \Upsilon_{\text{rob}} \textcolor{red}{\downarrow} \)                   & 0.206      & 0.063 & 0.185 & 0.246 & {\cellcolor{myred}\textbf{0.299}} & 0.097 & 0.191          \\
                       & \(\mathcal{E}_{\text{rate}} \textcolor{green}{\uparrow} \)                    & {\cellcolor{mysalmon} \textbf{-0.060}}       & -0.016 & {\cellcolor{mysalmon} \textbf{-0.057}} &\cellcolor{mysalmon}\textbf{-0.079} &\cellcolor{mysalmon}\textbf{-0.062} &\cellcolor{mysalmon}\textbf{-0.030} &\cellcolor{mysalmon}\textbf{-0.057}           \\
                       & \(\kappa_{\text{NIF}} \textcolor{green}{\uparrow}\)                    & 5.292 & 5.209 & 5.384 & 5.288 & 5.00 & 5.169       & 5.276          \\ \hline
                                   
\end{tabular}%
}

\caption{\label{citation-guide}
    This table showcases multiple performances of the models under diverse noise conditions, quantified through the Robustness Index, Error rate and Noise Impact Factor (NIF). \( \textcolor{green}{\uparrow} \) indicates higher is better while \( \textcolor{red}{\downarrow} \) indicates the opposite. \colorbox{mygreen}{\textbf{Green}},\colorbox{myblue}{\textbf{blue}} and \colorbox{mylavender}{\textbf{lavender}} indicate the best while \colorbox{myred}{\textbf{red}}, \colorbox{mysalmon}{\textbf{salmon}} and \colorbox{myyellow}{\textbf{yellow}} indicate the worst on robustness score for every metric under each of the adversarial noise.
  } 

\label{tab:metric performance}
\end{table*}

\section{Experimental Results}

\subsection{Levels of Noise}
As noise is introduced to the paragraph contexts and its severity increases, the accuracy of the model declines. Table \ref{tab:noise-performance} shows that the performance of each model varies depending on the level of noise severity, which is expected as higher noise levels complicate the ability of the model to produce accurate outputs. We applied noise at multiple severity levels across the character, word, and sentence levels, and observed that QA models are vulnerable to all types of noise. The primary comparisons are drawn from how the models perform as levels increase under a particular noise. 


\begin{table*}[ht]
\centering
\vspace{0.2cm}
    \begin{tabular}{ccc}
    \toprule
     \textbf{Ground Truth} & \textbf{DeBERTa Answer} & \textbf{DistilBERT Answer}\\
    
  \midrule
   The Mac Plus & Mac Ps & macinosh i aple\\
   Mac OS 8 & Mc O 8 & banded mc o 8\\
   high voltage & hg voltag & hg voltag\\
   MacWrite and MacPaint & MacWrite a acPaint & macwrite a acpaint \\
   GUIs in general & U i general & u i general \\

   \bottomrule
  
\end{tabular}

\caption{\label{Char_del_performance}
    Outputs of DeBERTa and DistilBERT for the noise "Char Del" at the severity level 5.
  } 

\label{tab:answerError}

\end{table*}

\subsection{Metrics Analysis}
\begin{figure*}[ht]
  \centering
  \includegraphics[width=\textwidth]{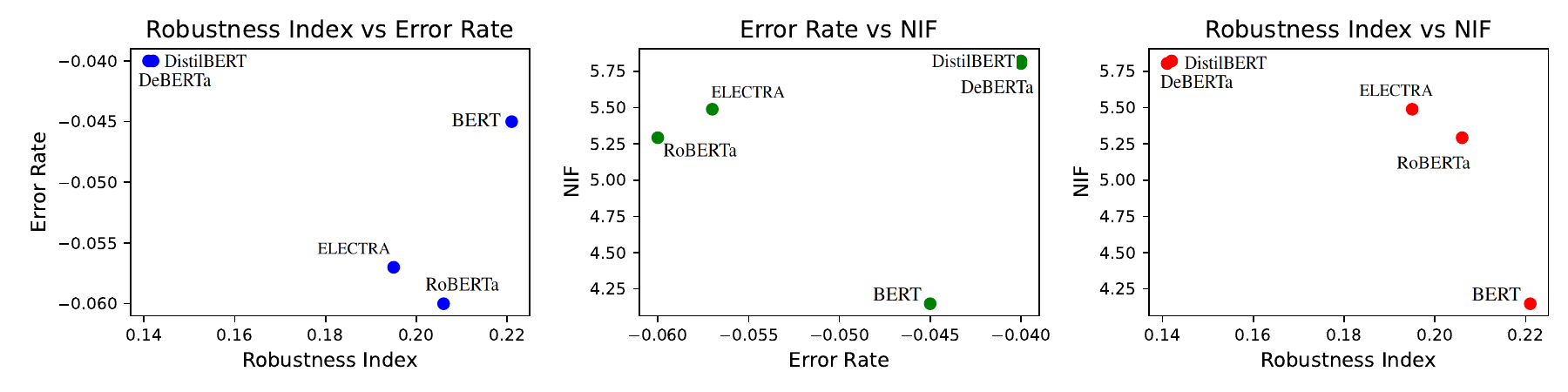}
  \caption{Scatter plots showing the relationships between different metrics under the noise "Char Del": Robustness Index vs Error Rate, Error Rate vs NIF, and Robustness Index vs NIF.}
  \label{fig:Metrics behave}
\end{figure*}

Our three robustness metrics—Robustness Index, Error Rate, and NIF— capture different aspects of the robustness of QA models. Table \ref{tab:metric performance} shows that the "Robustness Index" and "Noise Impact Factor" exhibit similar patterns, while the "Error Rate" ranks the models slightly different, as illustrated in Figure \ref{fig:Metrics behave}. The data we gathered clearly shows that the metrics yield different values for each model under varying noise conditions. Table \ref{tab:metric performance} provides insights into the strengths and weaknesses of each model across different noise levels for each metric, demonstrating that each metric offers perception insights into the performance of the models. 

\subsection{Analysis of Noise Impact}

\begin{figure}[ht]
  \includegraphics[width=0.875\columnwidth]{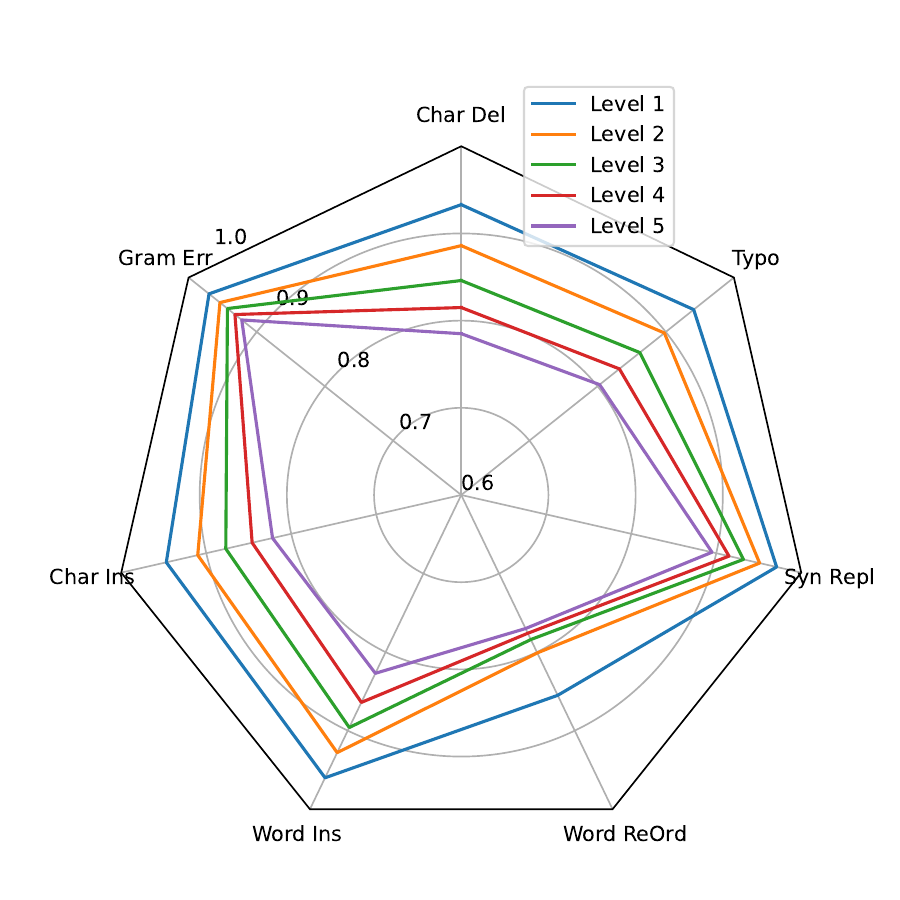}
  \caption{Performance of DeBERTa under Different Noise Types and Levels.}
  \label{fig:DeBERTa performance}
\end{figure}
 From Fig. \ref{fig:DeBERTa performance}, the most significant factors are "Character Deletion", "Typing Mistake", and "Word Reordering". With Fig. \ref{fig:DeBERTa performance}, we can infer that reordering words drastically alters the meaning of the context and often misleads QA models into providing incorrect answers. Word-level perturbations, such as "Character Deletion", frequently cause the model to select the wrong index or answer, resulting in lower accuracy, shown in Table \ref{tab:answerError}. On the other hand, "Character Insertion", "Synonym Replacement", and "Word Insertion" have a moderate impact, as the primary meaning of the context often remains intact, making it easier for the models to identify the correct answer. Finally, "Grammatical Errors", which we introduced in this research, have little to no effect in some cases.

\subsection{Model Evaluation}

DeBERTa and DistilBERT outperform other models across all types of noise, maintaining the highest levels of robustness with DistilBERT being 30.58\% more base accurate than BERT. From
DeBERTa leads the chart of the metrics by a small margin from DistilBERT, indicating that for estimating robustness, DeBERTa is superior. Both of these models consistently achieve top rankings for almost all of the noise across all metrics, suggesting that their architecture is better suited for maintaining contextual integrity. 

BERT, on the other hand, performs the worst. All of its base variants scores high in accuracy. However, Table \ref{tab:metric performance} shows that BERT outperforms both RoBERTa and ELECTRA in terms of "Error Rate", suggesting that its accuracy does not drop as sharply as in other models, resulting in a less steep slope and, consequently, a higher error rate. While "Word Reordering" significantly impacts the accuracy of most models, BERT appears to be resilient to increased severity levels, with its error rate only slightly lower than that of DistilBERT. 

While RoBERTa and ELECTRA do not perform as well as DistilBERT or DeBERTa, they still show solid results, notably outperforming BERT. ELECTRA surpasses RoBERTa, particularly excelling in the "Word Insertion" task, where it delivers surprisingly strong performance compared to all other model.

\subsection{Accuracy vs Evaluation Metrics}
Both Table \ref{tab:noise-performance} and Table \ref{tab:metric performance} provide valuable insights into the models’ behavior. While our evaluation metrics primarily focus on accuracy, they also offer a comprehensive view of model performance by accounting for all severity levels. As a result, a model with lower accuracy could still perform well in terms of robustness. Fig. \ref{fig:Metrics behave} is the representation of accuracy cannot define the robustness as "Error Rate" is higher for with lower accuracy models

\subsection{Additional Analysis}

From Table \ref{tab:noise-performance}, we observe a significant declination in the performance of the models when subjected to the two types of noise -- "Character Deletion" and "Word Reordering". Two key conclusions can be drawn from here. Firstly, when character-level perturbations occur, models need to either identify the correct word despite the perturbations or generate the proper word on their own. In our experiments, both approaches failed, as none of the models demonstrated the ability to reconstruct or recognize the correct word when characters were missing. This highlights a limitation in the models' ability to handle fine-grained character-level noise.

Secondly, semantic understanding appeared to be a critical requirement for models, as they struggled significantly when words within a sentence were shuffled. The models often became confused and were unable to provide accurate answers. If the models possessed a deeper grasp of sentence semantics, they would likely be more capable of identifying or predicting the correct answers, even when the order of the words was altered. This suggests that improving semantic reasoning could enhance the robustness of the model in dealing with noisy or perturbed contexts.

\section{Discussion}
The adaptability of the framework allows the integration of more models and adversarial noise. The implementation of applying noise can differ in two ways. The noisy dataset can be generated first and then the inferences can be performed. Alternatively, the noise can be applied to the textual input during runtime. The latter approach can serve as resource efficient alternative. While SQuAD is a widely recognized benchmark for contextual question answering, our framework is designed to be flexible and can integrate additional contextual QA datasets to address and minimize potential dataset biases.



\section{Conclusion}

Our work offers valuable insights into how QA models respond when exposed to noise in their input contexts. We have developed a novel evaluation framework to analyze the robustness of QA models by introducing various types of noise at different levels of severity. This approach allows others to evaluate the robustness of QA systems by using their own datasets. Our goal was to highlight specific areas where models can improve and identify the types of noise they should be trained on to enhance their robustness against real-world challenges.

\section*{Limitations}

Our research focused on $5$ widely used transformer-based models. Including the results of more such questions answering model can enrich the analysis. While the empirical evaluation of model robustness has been shown, the theoretical formulation of robustness in the task of contextual QA has not been explored.

\bibliography{custom}

\end{document}